\documentclass[sigconf]{acmart}

\setcopyright{none}
\renewcommand\footnotetextcopyrightpermission[1]{}
\acmConference[HRI '27]{Proceedings of the 22nd ACM/IEEE International Conference on Human-Robot Interaction}{March 8--12, 2027}{Santa Clara, CA, USA}
\acmYear{2027}
\copyrightyear{2027}

\usepackage{amsmath}
\usepackage{booktabs}
\usepackage{enumitem}
\usepackage{graphicx}
\usepackage{microtype}
\usepackage{xcolor}
\newcommand{\NTestCheckpoints}{162}
\newcommand{\NInterventions}{972}
\newcommand{\GAVAAccuracy}{100\%}

\newcommand{\NUnitTests}{72}

\hypersetup{
  pdftitle={Knowing When to Yield: Grounded Arbitration of User Corrections in Text-Based Embodied Agents},
  pdfauthor={Yezhou Cheng, Runjia Du, Zeming Liu, Hang Lyu, Zehua Yang, Bojun Lin},
  pdfsubject={HRI 2027 arXiv preprint},
  pdfkeywords={human-robot interaction, embodied agents, language correction, verification}
}

\title{Knowing When to Yield: Grounded Arbitration of User Corrections in Text-Based Embodied Agents}

\author{Yezhou Cheng}
\affiliation{%
  \institution{University of Wisconsin - Madison}
}
\email{cheng222@wisc.edu}

\author{Runjia Du}
\affiliation{%
  \institution{Independent Researcher}
}

\author{Zeming Liu}
\affiliation{%
  \institution{Brown University}
}

\author{Hang Lyu}
\affiliation{%
  \institution{Brown University}
}

\author{Zehua Yang}
\affiliation{%
  \institution{Independent Researcher}
}

\author{Bojun Lin}
\affiliation{%
  \institution{Pinterest Inc}
}

\renewcommand{\shortauthors}{Cheng et al.}

\begin{abstract}
How should an embodied agent respond when a person's correction may be wrong?
We formulate grounded correction arbitration as a choice among accepting,
rejecting, inspecting the world, and asking the speaker. GAVA implements this
interface with observation-bounded evidence, legal probes, and a one-step
expected-loss rule. We contribute an executable protocol and a controlled
account of checking, clarification, and prior knowledge.
In text-only ALFWorld, \NTestCheckpoints{} checkpoints produce
\NInterventions{} paired true/false interventions. Complete local inspections
give GAVA and always verify \GAVAAccuracy{} correction accuracy, establishing
the evidence contract rather than a comparative advantage. Same-episode
execution with an independent train-prior planner counts realized actions and
clarification exchanges. With a perfect speaker, GAVA reduces interaction cost
against always verify but ties a cost threshold. A locally frozen
fallible-speaker study provides only nominal evidence for reliability
calibration under a uniform prior.
An exploratory extension uses training-only object--location counts as a claim
prior. On 340 unseen scenarios it lowers interaction and declared joint cost
against uniform GAVA by $-0.490$ and $-0.420$. We then freeze the policy, costs,
matched baselines, and multiplicity plan before internal replication on 77
non-overlapping seen checkpoints (308 scenarios). The respective gains replicate:
$-0.595$ and $-0.517$, with both 95\% checkpoint-bootstrap CIs excluding zero.
Joint cost also improves against an identical-prior fixed policy ($-0.639$,
CI [$-1.275,-0.166$]). The identical-prior calibrated no-VOI comparison remains
inconclusive ($-0.208$, CI [$-0.630,0.224$]); it establishes neither superiority
nor equivalence. Semantic GAVA makes four factual errors in each cohort
(98.8\%/98.7\% accuracy). All methods complete every task.
The results support selective information gathering with semantic priors under
declared costs. They do not isolate a general benefit of environmental VOI over
clarification. The study uses normalized claims, complete symbolic observations,
and controlled speakers; no human participants, visual input, or physical robot
are evaluated.

\end{abstract}


\begin{document}
\maketitle
\hypersetup{
  pdftitle={Knowing When to Yield: Grounded Arbitration of User Corrections in Text-Based Embodied Agents},
  pdfauthor={Yezhou Cheng, Runjia Du, Zeming Liu, Hang Lyu, Zehua Yang, Bojun Lin},
  pdfsubject={HRI 2027 arXiv preprint},
  pdfkeywords={human-robot interaction, embodied agents, language correction, verification}
}

\section{Introduction}

Language offers an appealing interface for correcting a robot while it works.
Prior systems map utterances such as ``not there'' or ``do this instead'' into
plan edits, policy updates, or revised actions
\cite{thierauf2024dothis,sharma2022correcting,cui2023lilac,liu2023olaf,shi2024yell}.
This literature largely studies whether a robot can \emph{use} a correction.
It usually does not test whether the correction should be trusted.  That
assumption is consequential: people can misremember an object location, refer
to the wrong receptacle, or propose an action whose preconditions are not yet
known.  A robot that treats every correction as supervision may exchange one
planning error for another.

Language-model research calls uncritical agreement with a user's false premise
\emph{sycophancy}, and distinguishes it from stubborn rejection of a valid
correction \cite{sinha2026sycobench,pi2025pointing,beigi2025smart}.  A recent HRI
scoping review frames robotic sycophancy as an emerging interaction concern and
calls for dedicated measurement \cite{seaborn2026robotic}. Existing
benchmarks, however, mostly score an answer to a static question.  An embodied
agent can act to obtain evidence, and its choice has step, dialogue, and physical
consequences.  Thus the relevant question is not simply whether the model agrees;
it is whether the agent chooses an appropriate, grounded response before changing
the world.

We study this question without visual rendering or human participants.  The
text-only ALFWorld interface aligns symbolic household tasks with an interactive
language environment \cite{shridhar2021alfworld}.  It exposes legal actions and
text observations while retaining hidden state for evaluation, allowing a
strict separation between what a deployable method observes and what the
experimenter uses to label a correction.

This paper contributes:
\begin{enumerate}[leftmargin=*,nosep]
  \item a paired protocol for factual corrections at matched early checkpoints
  across six task families. Observable inputs are separated from hidden labels;
  three wording styles check the normalized-claim interface;
  \item GAVA (Grounded Advice Verification and Arbitration), an auditable
  one-step rule for \textsc{accept}, \textsc{reject}, \textsc{verify}, and
  \textsc{clarify}. Its complete-listing contract makes positive and
  claim-local negative evidence executable; and
  \item a controlled comparison separating evidence resolution from interaction
  efficiency: static ablations test the observation contract; online runs count
  actions and dialogue; an exploratory semantic prior and a prospectively frozen
  internal replication test cost savings against matched policies.
\end{enumerate}
Semantic priors reduce checking relative to uniform GAVA and always verify;
the full policy lowers declared joint cost relative to prior-only decisions.
Added value over matched calibrated clarification remains uncertain. The HRI
contribution is a correction-handling interface and evaluation protocol.
Simulated users supply controlled interventions; the study does not establish
human trust, natural dialogue competence, or physical safety.

\section{Related Work}

\subsection{Language correction in robotics}

Natural-language feedback has been used to revise robot cost functions
\cite{sharma2022correcting}, alter actions and goals
\cite{thierauf2024dothis}, update policies \cite{liu2023olaf,shi2024yell}, and
provide shared-autonomy corrections \cite{cui2023lilac}. These systems
establish that verbal correction can be useful, but typically treat it
as task-relevant supervision.  Thierauf et al. explicitly identify unsuitable
or invalid task-level corrections and uncertainty about the desired response as
open limitations \cite{thierauf2024dothis}.  GAVA addresses that boundary: it
arbitrates factual/procedural claims before a downstream learner or planner uses
them.

\subsection{Grounded feedback and closed-loop planning}

Embodied language agents improve when they feed environment observations back
into planning \cite{huang2022inner,bhat2024brainbody}.  Those feedback channels
report execution progress or state; they are not conversational claims that may
be true or false.  We combine closed-loop state feedback with an explicit source
distinction: observations update the belief store, whereas a correction remains
a hypothesis until entailed, contradicted, or actively checked.

\subsection{Active query and clarification}

Information-seeking interaction is established in HRI. Deits et al. select
clarifying questions by expected entropy reduction and merge human answers into
a grounded command model \cite{deits2013clarifying}. Cakmak and Thomaz show that
robot query types impose different human costs and preferences
\cite{cakmak2012questions}; grounded clarification has also been evaluated end
to end.
Text agents can also learn when to query an external knowledge source
\cite{liu2022afk}, while ReSpAct interleaves reasoning, dialogue, and action in
ALFWorld \cite{dongre2025respact}, and SECURE uses embodied conversation to
acquire previously missing concepts \cite{rubavicius2026secure}.
Thus neither VOI nor clarification is our novelty.
The distinct object here is a correction whose factual content may itself be
wrong, and the choice between querying the world and querying the speaker:
GAVA checks that source before using it as supervision, while the benchmark
pairs true/false claims and isolates runtime observations from label state.

\subsection{Sycophancy and correction selectivity}

SycoBench-600 measures whether language models accept correct feedback and resist
incorrect feedback \cite{sinha2026sycobench}; related work exposes a
sycophancy--stubbornness trade-off in visual question answering
\cite{pi2025pointing} and uses uncertainty-aware reasoning to mitigate textual
sycophancy \cite{beigi2025smart}.  Our work does not claim to originate
\emph{correction selectivity}.  It operationalizes selectivity in executable,
partially observed household tasks, adds legal evidence-gathering actions and
their cost, and evaluates state-grounded consequences rather than answer text.

Table~\ref{tab:related-work} makes the resulting gap explicit. Robot-correction
work supplies the interaction setting but normally assumes useful guidance;
sycophancy benchmarks vary feedback truth but lack actions; closed-loop planners
act on observations but do not arbitrate a fallible conversational source. GAVA
is intended as a bridge among these capabilities, not a replacement for the
learning or planning methods in the first two rows.

\begin{table*}[t]
  \caption{Relationship to adjacent research. ``Yes'' means the capability is
  central to the cited evaluation, not merely possible in principle.}
  \Description{A comparison table showing that prior robot correction,
  closed-loop planning, and sycophancy benchmarks each cover only subsets of
  executable action, fallible user feedback, active verification, and explicit
  observable-state isolation; this work covers all four in a text simulator.}
  \label{tab:related-work}
  \centering
  \small
  \begin{tabular}{p{0.31\textwidth}cccc}
    \toprule
    Research line & Executable actions & Fallible correction & Active probe & Observable-state boundary \\
    \midrule
    Robot language correction
      \cite{thierauf2024dothis,sharma2022correcting,cui2023lilac,liu2023olaf,shi2024yell}
      & Yes & -- & -- & varies \\
    Closed-loop embodied LLM planning
      \cite{huang2022inner,bhat2024brainbody}
      & Yes & -- & feedback & varies \\
    Active HRI query and clarification
      \cite{deits2013clarifying,cakmak2012questions}
      & Yes & -- & dialogue & Yes \\
    Text/visual correction selectivity
      \cite{sinha2026sycobench,pi2025pointing,beigi2025smart}
      & -- & Yes & -- & -- \\
    This work & Yes & Yes & Yes & Yes \\
    \bottomrule
  \end{tabular}
\end{table*}

\section{Grounded Correction Arbitration}

Figure~\ref{fig:pipeline} summarizes the information boundary. The method-facing
API omits the evaluation record, and the scorer never participates in action
selection. In integrated runs, a separate evaluation process owns the oracle;
policy workers hold neither the label record nor the simulated-user object.
Their only reply channel is a one-shot capability bound to a specific
clarification event. This protects the audited trusted-code execution path; it
does not claim filesystem or OS sandboxing of adversarial method code.

\begin{figure*}[t]
  \centering
  \fbox{\parbox{0.96\textwidth}{\centering
  \textbf{Public runtime:}\quad correction $c_t$
  $\rightarrow$ atomic claims $P(c_t)$
  $\rightarrow$ observable belief $B_t$
  $\rightarrow$ evidence \{entailed, contradicted, unknown\}
  $\rightarrow$ expected-loss/VOI arbitration
  $\rightarrow$ accept, reject, legal probe, or clarify\\[3pt]
  \textbf{Legal feedback loop:}\quad probe action
  $\rightarrow$ TextWorld observation
  $\rightarrow B_t\oplus o$
  $\rightarrow$ re-arbitration\\[3pt]
  \textbf{Evaluation-only path (separate file/process):}\quad hidden facts
  $\rightarrow$ private twin labels $\rightarrow$ post-clarify text service
  and post-hoc scoring}}
  \caption{GAVA pipeline and audited trusted-code oracle boundary. Hidden facts
  generate and score interventions but are absent from method inputs.}
  \Description{A three-line flow diagram. Public corrections become atomic
  claims, observable evidence, and arbitration decisions. Legal actions return
  text observations to the belief state. A separate evaluation path uses hidden
  facts only for labels and scoring.}
  \label{fig:pipeline}
\end{figure*}

At step $t$, an agent has goal $g$, observation history $h_t$, and an
observation-bounded belief state $B_t$. The interface represents correction
$c_t$ as atomic propositions $P(c_t)=\{p_1,\ldots,p_k\}$ and optional edits.
The benchmark supplies normalized claims; learned parsing is not evaluated.
For each $p_i$, the verifier returns
\begin{equation}
e_i \in \{\textsc{entailed},\textsc{contradicted},\textsc{unknown}\}.
\end{equation}
Missing evidence is \textsc{unknown}, not false. Functional predicates add only
ontology-licensed contradictions. We do not declare ALFWorld's $at$ relation
globally functional: co-located receptacle names can both list the same object.
A location claim is contradicted only by a complete local listing of the claimed
receptacle that omits the object.

The response set is
$D=\{\textsc{accept},\textsc{reject},\textsc{verify},\textsc{clarify}\}$.
For an unknown claim with model prior $\pi=\Pr(p_i\mid B_t)$, GAVA compares
one-step expected asymmetric loss within this response class:
\begin{align}
R_A&=(1-\pi)C_{FA}, &R_R&=\pi C_{TR},\\
R_F&=\min\{R_A,R_R\},\\
R_{CL}&=C_{CL}+(1-\gamma)R_F\notag\\[-2pt]
&\quad+\gamma(1-\eta)[\pi C_{TR}+(1-\pi)C_{FA}],\label{eq:clarification}
\end{align}
where $C_{FA}$ and $C_{TR}$ penalize false acceptance and true rejection;
$C_{CL}$ is clarification cost. Resolution probability $\gamma$ and conditional
answer accuracy $\eta$ are assumed for either truth value.
Equation~\eqref{eq:clarification} evaluates a restricted policy: follow a
resolved reply; otherwise use the best fixed response. It does not perform
a posterior Bayes decision on the reply. Let $r_0=\min\{R_A,R_R,R_{CL}\}$.
A public probe $q$ has a legal macro, cost $C(q)$, and probability $\rho_q$ of
complete, truthful resolution. Non-resolution is modeled as leaving the prior
unchanged and incurring fallback risk $r_0$. The one-step probe risk and value are

\begin{equation}
R_q=C(q)+(1-\rho_q)r_0,
\qquad \mathrm{VOI}(q)=r_0-R_q.\label{eq:probe}
\end{equation}
GAVA selects the $q$ with minimum $R_q$ only when
$\mathrm{VOI}(q)>\tau$; otherwise it returns the minimum-risk non-probe
decision. Equal fixed risks favor accept; equality between clarification and
the fixed fallback favors the fixed response. In this study
$\tau=0$. Static response selection assumes $\gamma=\eta=1$ but stops at a
clarification request (Section~\ref{sec:heterogeneous}); integrated runs execute
the reply. The frozen reliability study varies both parameters by checkpoint.
Uniform-prior runs set $\pi=.5$. The exploratory semantic variant sets $\pi$ within the exposed pair
to $(n(x,r)+1)/(n(x,r)+n(x,r')+2)$, where $n$ is the frozen ALFWorld-training
object--location count and $r'$ is the public symmetric mate; it reads no test
label. This is a smoothed, pair-conditioned estimate; ranking accuracy does
not establish probability calibration. Entailed and contradicted evidence is hard under the benchmark's
complete-listing contract ($\pi=1$ and $0$, respectively), so its residual
fixed-decision loss is zero. A noisy deployment would replace this contract
with a calibrated observation model $P(o\mid B_t,q)$ and the full expectation;
that extension is not evaluated here.

\subsection{Decision procedure}

GAVA queries $B_t$: entailed claims are accepted and contradicted claims
rejected without action. Unknown claims use the best fallback over accept,
reject, and clarify. A location probe requires admissible \texttt{go to r}.
Its macro navigates to $r$ and opens it only if \texttt{open r} is then
admissible. Complete listings license $\rho_q=1$ from public commands and
observation semantics, without hidden facts.

For complete probes, Equation~\eqref{eq:probe} gives $R_q=C(q)$: inspection is
chosen exactly when its cost is below the best non-probe risk. With symmetric
loss $C_{FA}=C_{TR}=C$, this condition becomes
\begin{equation}
C(q)<\min\{C\min(\pi,1-\pi),R_{CL}\}.
\label{eq:threshold}
\end{equation}
This explains the perfect-speaker cost-threshold tie. Semantic priors and
fallible replies alter the fallback risk, not the decision rule.

Figure~\ref{fig:algorithm} shows complete-listing execution; only returned
observations update $B_t$. Incomplete descriptors use $\rho_q<1$ and
Equation~\eqref{eq:probe}'s one-step fallback; repeated unresolved probing is
not evaluated. This local risk omits downstream search costs and does not
guarantee minimum episode cost.

\begin{figure}[t]
  \centering
  \fbox{\parbox{0.93\columnwidth}{\small
  \textbf{Input:} correction claims $P(c_t)$, observable belief $B_t$,
  admissible actions $A_t$, and costs.\\
  \textbf{For each claim $p$:}
  \begin{enumerate}[leftmargin=1.5em,nosep]
    \item Query $e\leftarrow B_t(p)$.
    \item If entailed, accept; if contradicted, reject.
    \item Otherwise compute the minimum-risk non-probe fallback over
      accept, reject, and clarify.
    \item Construct probes $q$ from admissible commands $A_t$ that can reveal $p$.
    \item If the best probe has strictly positive VOI, execute it, update
      $B_t$ only from its returned observation, and go to step 1.
    \item Otherwise return the non-probe fallback.
  \end{enumerate}}}
  \caption{GAVA under complete-listing probes ($\rho_q=1$).
  Evaluation-only facts are not inputs.}
  \Description{Pseudocode that queries observable evidence, handles entailed
  and contradicted claims, compares unknown-claim fallbacks with legal probes,
  updates only from returned observations, and repeats.}
  \label{fig:algorithm}
\end{figure}

\subsection{Local negative evidence}

Open-world reasoning creates a subtle failure mode: observing a listed object is
positive evidence, but global absence is not evidence of falsehood.  ALFWorld
nevertheless returns a complete visible-content enumeration after reaching an
open surface or opening a container.  GAVA therefore licenses a negative
$\neg at(x,r)$ only when (1) the agent legally inspected $r$, (2) the environment
returned a complete local listing, and (3) $x$ was absent.  It does not infer
that an unmentioned object is absent elsewhere.  The implementation also binds
the tightly scoped anaphor ``in it'' only to a container confirmed as opened in
the same observation. It likewise does not infer $\neg at(x,r_2)$ merely from
seeing $x$ at $r_1$, because ALFWorld can expose one object through multiple
co-located receptacle names.

\subsection{Claim-level decisions}

Decisions are made per proposition.  A multi-claim correction can therefore be
partially accepted: supported claims enter the planner, contradicted claims are
rejected, and unknown claims remain pending.  The present confirmatory benchmark
uses one controlled location claim per intervention to isolate arbitration from
semantic parsing; multi-claim behavior is covered by executable unit tests and
left for a broader benchmark.

\subsection{Worked trace}

One unseen task asks the agent to place a saltshaker in a drawer. The simulated
correction claims that \texttt{saltshaker 1} is at \texttt{cabinet 2}. Initially
the proposition is unknown and both accepting and rejecting have expected loss
0.5. With clarification cost 0.25 and normalized inspection-macro cost 0.10,
GAVA selects
\textsc{verify} (VOI 0.15). The first legal action reports that the cabinet is
closed, which is not location evidence. GAVA opens it; the complete response is
``In it, you see a glassbottle 1, and a plate 2.'' The contextual ``it'' is bound
to the cabinet confirmed open in that same observation. Because the complete
local list omits the saltshaker, the claim becomes contradicted and is rejected.
At no point does the runtime access the oracle's true location.

\section{Paired Embodied-Correction Benchmark}

\subsection{Environment and tasks}

We use ALFWorld's TextWorld interface, not its Unity renderer
\cite{cote2018textworld}.  The six task
families require locating, moving, examining, cleaning, cooling, heating, or
placing household objects.  Each record stores the goal, observation, legal
commands, replay prefix, correction text, and normalized claim.  The method
never receives simulator predicates or the expert policy.

\subsection{Intervention construction}

For each accepted game, the evaluation-only generator reads the initial expert
policy and hidden facts. If the policy begins by navigating to and taking a
target object from a factually validated receptacle, it creates a true location
claim and a minimally different false twin naming another currently reachable
receptacle. To prevent the false-member construction from encoding the label,
all legal locations are hash-ranked from public checkpoint fields and paired by
an adjacent, fixed-point-free involution: if $A$ maps to $B$, the same public
rule maps $B$ to $A$. With an odd number of locations, the last location is
unmatched; a checkpoint whose true location is unmatched is excluded rather
than given a directional fallback. Both twins share the goal, checkpoint,
object, legal-action set, and wording template. We render neutral, doubt, and
authority styles. Templates change pragmatic pressure but not propositional
content.

Pilot games are excluded from the unseen test set by path.  We evaluate separate
ALFWorld \texttt{valid\_seen} and \texttt{valid\_unseen} splits; these names
describe ALFWorld's learned-agent split, not newly sampled domains. GAVA fits
no policy, but its semantic variant estimates a prior from training counts. Invalid cases
are rejected before any method output is inspected: short policies,
non-location openings, policy/fact mismatches, absence of a legal false
location, or an unmatched true location. Checkpoint, pair, and skip counts plus file hashes are saved in
machine-readable manifests.

Table~\ref{tab:selection} reports the complete selection flow. All 134 unseen
games were enumerated; after removing 26 pilot games, all remaining 108 were
inspected. For the secondary seen analysis, predeclared per-family quotas caused
inspection to stop after 111 of 140 games, leaving 29 uninspected. The 21 seen
timeouts, rather than being scored as failures, are outside the selected stratum
and are a stated validity threat.

\begin{table}[t]
  \caption{Checkpoint selection flow. ``Other'' includes non-location policy
  openings, policy/fact mismatches, and unmatched locations.}
  \Description{The complete flow from all candidate games through pilot or
  quota exclusions, inspection, rejection, and final accepted checkpoints.}
  \label{tab:selection}
  \centering
  \scriptsize
  \begin{tabular}{lrrrrrr}
    \toprule
    Split & All & Pilot/quota & Inspected & Timeout & Other & Accepted \\
    \midrule
    Unseen & 134 & 26 & 108 & 0 & 23 & 85 \\
    Seen & 140 & 29 & 111 & 21 & 13 & 77 \\
    \bottomrule
  \end{tabular}
\end{table}

\begin{table}[t]
  \caption{Frozen benchmark composition. ``Pairs'' counts pressure-specific
  true/false twins; the statistical unit is the checkpoint.}
  \Description{The unseen split contains 85 checkpoints, 255 true-false pairs,
  and 510 interventions. The seen split contains 77 checkpoints, 231 pairs, and
  462 interventions.}
  \label{tab:dataset}
  \centering
  \small
  \begin{tabular}{lrrr}
    \toprule
    Split & Checkpoints & Pairs & Interventions \\
    \midrule
    Valid unseen & 85 & 255 & 510 \\
    Valid seen & 77 & 231 & 462 \\
    \textbf{Total} & \textbf{162} & \textbf{486} & \textbf{972} \\
    \bottomrule
  \end{tabular}
\end{table}

\subsection{Why the twins remain partially observed}

The intervention is injected before the base policy's first navigation action.
The reset observation enumerates reachable furniture but not the target's
location. Consequently, both the true and false member begin as unknown to the
runtime method even though the generator can validate them against hidden facts.
This design rules out a trivial strategy that simply restates an already observed
location. It creates the same observation-state obligation for both twins, but
does not assume equal semantic plausibility: training-world knowledge may favor
one member, which we audit and use in a claim-adaptive extension. Once a complete local listing is returned, positive
and claim-bounded negative evidence are equally actionable.

\subsection{Oracle isolation and leakage controls}

Public and oracle JSONL files are physically separate.  Public records contain
the claim but no truth label, true location, hidden fact list, or expert policy.
Their IDs are 24-hex hashes of public content, and rows are sorted by opaque ID
rather than private twin order. Tests over both released splits forbid
truth-bearing keys/tokens and enforce the format: all 2,916 ID fields pass, and
the lexicographically first member is true in 250/486 pairs. Every released
pair passes the public involution in both directions, so either member is
construction-consistent as the true claim. As negative controls, the former
minimum-location rule scores 50.6\% unseen and 49.4\% seen, while a deterministic
public pair-hash guess scores 51.4\% and 49.8\%; all Wilson intervals include
50\%. These checks rule out the identified deterministic construction decoder,
not semantic inference from ordinary task knowledge. Runners accept only the
public path; evaluation code alone accepts both files. Tests also verify that
the adapter requests observations and legal commands, not simulator facts.
For integrated clarification, a dedicated service process owns oracle and
simulated-user state. Policy workers hold only authenticated transport data and
one scenario/claim-bound, one-shot capability. Simulated-user state remains
server-side; unknown, reused, premature, or mismatched capabilities are
rejected. The service is invoked lazily after choosing \textsc{clarify}. This is
a hardened audited boundary for trusted method code, not an OS sandbox against
malicious code or direct access to installed ALFWorld files.

\section{Evaluation}

\subsection{Research questions and outcomes}

We ask: (RQ1) does grounded arbitration resolve paired true/false claims;
(RQ2) which observation and probe mechanisms enable resolution; (RQ3) does
the rule choose lower model risk across heterogeneous evidence, probes,
reliability, cost, and loss; (RQ4) how do responses affect task success and
action cost; and (RQ5) can a training-only semantic prior reduce checking,
replicate internally, and improve on policies with the same prior but no
environmental VOI?

The static primary outcome is accuracy among resolved accept/reject responses:
accept true claims or reject false ones. Coverage is separate. A clarification
request is pending, not an answer. Online runs execute probes and replies before
scoring correctness. Other outcomes are false acceptance, false rejection,
their union as incorrect fixed-decision (IFD) rate, legal actions, extra steps,
model risk, and task success.

For perfect-speaker RQ4, the primary comparison is realized interaction cost:
actions plus cost-weighted, counted clarification exchanges. The locally frozen
fallible-speaker study uses declared joint cost: interaction cost plus realized
false-accept or true-reject loss. These are declared decision units, not measured
harm. Components and correctness are reported separately; eventual task recovery
can conceal correction errors.

Method comparisons measure uncertainty at the base checkpoint, averaging its
twins and timings. Wording styles check the normalized-proposition interface,
not human persuasion.

\subsection{Systems and ablations}

We compare full GAVA with always accept, always reject, always verify, and a
feasibility-only policy that accepts any correction whose proposed navigation is
legal; the stress test also uses the minimum-risk fixed-only fallback, including
clarification. We ablate complete-listing negative evidence and automatic
opening of a closed receptacle. A cost sweep varies $C(q)$ around the VOI
threshold. A public-input-only component stress test instantiates 11 declared
decision contexts at each checkpoint, including already known evidence, absent
or unreliable probes, competing probes, the VOI boundary, and asymmetric losses.

For RQ4, we use a deterministic reactive planner that receives only the public
goal, normalized target object, current text observation, and admissible
commands. Its initial search order uses object--receptacle counts estimated from
6,353 ALFWorld training trajectories; held-out expert policies and oracle facts
are unavailable during execution. After every legal action it replans from the
returned text, searches the next receptacle after a failed hypothesis, and
implements all six task families. This deliberately transparent planner isolates
the downstream effect of arbitration; it is not presented as a learned or
frontier LLM planner. In the integrated condition, arbitration runs online in
the same TextWorld episode either before planning or after the first
train-prior navigation action. Accepting adds the claimed location as a search
preference, rejecting adds none, verifying executes the currently legal
navigation/opening macro and re-arbitrates from its text, and clarification
returns one controlled text reply whose parsed answer updates the preference.
The reliability study additionally compares calibrated GAVA with the same
policy assuming a perfect speaker, a calibrated policy without environmental
VOI, and the grounded raw-cost threshold.
The claim-adaptive extension changes only $\pi$ using the frozen train prior.
After its exploratory unseen run, we freeze a non-overlapping seen replication
with two matched baselines: prior-only fixed Bayes choice and calibrated
accept/reject/clarify without environmental VOI. All methods run online; four
primitive arms additionally validate a scenario-complete policy replay.

\subsection{Implementation and analysis}

All correction decisions use deterministic rules. Uniform GAVA uses
$C_{FA}=C_{TR}=1$, prior $\Pr(p)=0.5$ when unknown, clarification cost $0.25$,
and normalized inspection-macro cost $0.10$ unless varied. This macro may contain
a navigation action and, for a closed receptacle, one opening action; actual
environment steps are reported separately, and the stress test varies resolution
probability and loss. Entailed and contradicted propositions are hard evidence
under the complete-listing contract. The integrated experiment instead uses an
environment-action cost of 1, clarification cost 1.75, and
$C_{FA}=C_{TR}=6$. A fixed SHA-256-ranked sample of 100 training games (1,199
legal receptacle probes) estimates expected probe length by receptacle type;
held-out execution records the realized one- or two-action macro. This makes a
surface check cheaper than dialogue while the expected cost of a closed/openable
container can exceed it. These are declared interaction units, not estimates of
human time or physical risk. We locally froze the reliability design before
generating its outcomes; it was not preregistered or externally timestamped. We
SHA-256-ranked the original 89 unseen checkpoints into a fixed $3\times3$ design with
$C_{CL}\in\{0.75,1.75,3.0\}$ and low/medium/high $(\gamma,\eta)$ values of
$(0.45,0.80)$, $(0.70,0.90)$, and $(0.95,0.98)$. Cell assignment used only
public checkpoint paths. After the symmetric-pair repair, we removed the four
ineligible checkpoints without reassigning any survivor; the final cells contain
8--10 checkpoints. Deterministic
hashes of the public, truth-independent pair ID instantiate whether a lazily
queried reply resolves and is correct, coupling both twins without exposing
which is true. An unresolved reply takes the same minimum-risk fixed fallback used in
$R_{CL}$. These parameters are controlled stress levels, not human estimates.
``Calibrated'' means supplied speaker reliability here, not calibration learned
from people or calibration of the semantic prior.
The test suite contains
\NUnitTests{} tests covering open-world semantics, local anaphora, arbitration,
record validation, lazy reply gating, action legality, planner state continuity,
and oracle leakage. We report paired checkpoint bootstrap intervals for method differences
and do not treat the three templates as independent samples. The initial
semantic extension and its sweeps remain post-review exploratory. Before new
seen policy outcomes for this extension, we froze its 77-checkpoint internal replication, matched
baselines, six comparisons, and Holm family; this is prospective only for those
outcomes, not a preregistration or new domain. Closed-loop
comparisons average the true/false twins---and, for integrated runs, both
intervention timings---within each checkpoint before 10,000 bootstrap draws and
a 20,000-draw paired sign-flip test (seed 20270909).
Scenario accuracy is reported as error counts and descriptive percentages:
twins and timings are correlated, so scenario-level binomial intervals are not
used as uncertainty estimates. Prior-ranking Wilson intervals use one pair per
checkpoint.

The main runtime uses Python 3.11.14 (x86\_64 under macOS), ALFWorld 0.4.2,
TextWorld 1.7.0, and a project-local data directory. Seeds, record hashes,
validation reasons, and raw action/dialogue traces are saved. The statistical
script uses 10,000 checkpoint bootstrap draws with seed 20270918.

\paragraph{Baseline semantics.}
Always accept and always reject resolve without inspecting. Feasibility-only
accepts if the correction's proposed navigation is currently admissible; because
both twins name reachable receptacles, it tests whether executability alone can
establish truth. No correction follows the train-prior search order in
closed-loop evaluation. Always accept makes the claimed location the first
search preference; always reject ignores it. Always verify follows the same
legal information-gathering trajectory whenever one is available but does not
compare its cost to dialogue. Always clarify requests the controlled reply in
every case. A transparent grounded cost heuristic verifies iff the
train-estimated probe length is below clarification cost; under the present
hard-evidence, symmetric-loss contract it is the closed-form special case of
GAVA and is expected to tie it when the speaker is perfect. In the reliability
study, uncalibrated GAVA likewise assumes $\gamma=\eta=1$; calibrated no-VOI
receives the declared $(\gamma,\eta)$ but cannot inspect the environment.

\section{Results}

\subsection{Dataset yield and validation}

The frozen evaluation contains 85 unseen and 77 seen checkpoints, or 162
independent checkpoints total. Each checkpoint produces a true/false twin under
three wording styles, yielding 972 interventions. The unseen generator
enumerated 134 games, excluded the 26-game pilot, and rejected 19 of the 108
inspected candidates; four otherwise eligible checkpoints were excluded by the
symmetric matching rule. The seen generator inspected 111 of 140 games under
predeclared per-family quotas: 21 exceeded the 10-second bound, nine had a
non-location opening, three had policy/fact mismatches, and one eligible
checkpoint was unmatched; 29 were uninspected
after quotas filled. Independent reconstruction produced byte-identical public
and oracle files. All \NUnitTests{} tests passed.

\subsection{Correction resolution}

Static coverage counts terminal accept/reject responses; accuracy is conditional
on coverage. Clarification requests remain pending. Full GAVA and always verify
achieve 100\% coverage and accuracy with 0\% incorrect fixed decisions. GAVA uses
1.14/1.23 probe actions per unseen/seen intervention. Always accept, always reject,
and feasibility-only each have 50\% accuracy and 50\% incorrect decisions: twins
have opposite truth but both locations are reachable.

Without local negative evidence, true claims are accepted after observation but
no false claim is rejected. Coverage is 50\%, with every resolved response correct.
Without opening containers, coverage falls to 85.9\% unseen (checkpoint-bootstrap
95\% CI [80.6, 90.6]) and 77.3\% seen ([71.4, 82.5]); resolved decisions remain
correct. Accuracy alone would conceal this non-resolution.

These ablations test the observation contract. Complete local evidence permits
resolution; removing it reduces coverage. Feasibility alone provides no truth
evidence. At the default low cost, GAVA ties always verify because every unknown
claim is legally inspectable. The experiment does not test noisy perception or
natural-language understanding.

\subsection{Heterogeneous arbitration}
\label{sec:heterogeneous}

With symmetric loss, $\pi=.5$, and clarification cost 0.25, GAVA verifies at
probe costs 0.00--0.24. At 0.25 or above it selects \textsc{clarify}.
The static harness stops at this request: coverage and incorrect fixed decisions
are zero. Equation~\eqref{eq:clarification} prices a subsequent answer that is
not executed here. This tests response selection, not realized dialogue loss.
The following online experiments execute and score replies.

The 11 public descriptors per checkpoint (1,782 decisions) vary evidence, probe
availability, reliability, cost, the VOI boundary, and asymmetric loss. Mean
model risk is 0.176 for GAVA, 0.192 for always verify, and 0.205 for the best fixed
fallback. Losses use the same model that selects responses. This checks the rule
on designed cases, rather than independently validating task gains or
probability calibration.

\subsection{Integrated online arbitration and planning}

The online experiment invokes arbitration in the same TextWorld
episode as the independent planner.
Every neutral true/false twin is run at both initial and post-first-navigation
timings: 340 unseen and 308 seen scenarios per method. The 100-game training
sample yields 1,199 resolved probe macros. On held-out always-verify runs, 266/48
unseen probes and 212/70 seen probes take one/two actions; three unseen unknown
claims have no currently legal probe. Thus costs and availability vary through
environment interaction rather than declared stress-test descriptors.

The perfectly informative-speaker boundary first establishes online plumbing.
GAVA lowers mean interaction cost relative to always verify by $-0.144$ unseen
(95\% CI [$-0.278,-0.028$]) and $-0.367$ seen
([$-0.714,-0.156$]), while completing every task without invalid actions. It
nevertheless ties the grounded cost threshold exactly on both splits. Its
unseen comparison with always accept is uncertain ($+0.018$, CI
[$-0.149,0.174$]). Final-response mutations change planner preference and
trajectory in 12/12 unseen and 6/6 seen cases, excluding a logging-only explanation.

\subsection{Frozen fallible-speaker evaluation}

\begin{figure*}[t]
  \centering
  \includegraphics[width=\textwidth]{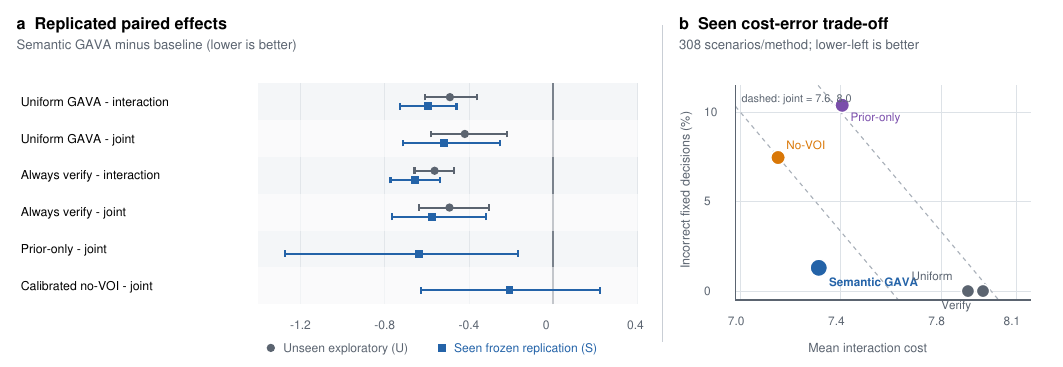}
  \caption{Semantic arbitration results. (a) Paired mean differences and 95\%
  checkpoint-bootstrap CIs; negative values favor Semantic GAVA. U is the unseen
  exploratory cohort and S the prospectively frozen seen replication. Matched
  baselines appear only in S. (b) Seen interaction cost versus incorrect
  fixed-decision (IFD) rate. Dashed contours show equal declared joint cost. All
  methods complete every task.}
  \Description{Panel a is a forest plot. Semantic GAVA has lower interaction and
  declared joint cost than uniform GAVA and always verify on both cohorts, and
  lower declared joint cost than prior-only on the seen replication. Its interval relative to
  calibrated no-VOI crosses zero. Panel b plots interaction cost against the
  percentage of incorrect fixed decisions. Semantic GAVA lies near the lower-left
  corner; no-VOI uses slightly less interaction but makes more factual errors.}
  \label{fig:semantic-results}
\end{figure*}

The locally pre-outcome-frozen reliability study assigns each unseen checkpoint to
one of nine dialogue-cost/informativeness cells without reading oracle labels or
method outputs. All eight methods run on the same 340 paired scenarios, yielding
2,720 main outcomes. Every method again has 100\% task success and zero invalid
actions. Interaction cost alone cannot distinguish a cheap but credulous response.
Figure~\ref{fig:semantic-results} summarizes the replicated effects and the
seen-split cost--error frontier for the subsequent semantic-prior extension.

Uniform-prior GAVA initially accepts 21 cases, clarifies 10, rejects three,
and verifies 306; after evidence or dialogue it makes 170 accepts, six corrected
responses, and 164 rejections, all factually correct. Uncalibrated GAVA and the
raw cost heuristic coincide, but make 7.6\% incorrect fixed decisions. Uniform-prior GAVA's
delta in declared joint cost against either is $-0.412$ (CI [$-0.793,-0.029$], raw
$p=.035$), but 48/85 checkpoints tie and the four-comparison Holm-adjusted
$p=.104$ makes this nominal evidence. Interaction cost is indistinguishable
($+0.047$, CI [$-0.168,0.313$]).

Versus calibrated no-VOI, uniform-prior GAVA's delta in declared joint cost is
$-1.796$ (CI [$-2.150,-1.393$]), supporting environmental evidence within the
uniform-prior setting.
Against always verify, it is $-0.073$ (CI [$-0.174,-0.004$]). Although this
checkpoint-bootstrap interval for the mean excludes zero, the
predeclared paired sign-flip test does not reject the paired null ($p=.122$);
with 81/85 ties, we therefore do not treat this as robust evidence of broad dominance.

The high-information, cost-0.75 cell favors the uncalibrated policy by 0.863
units of declared joint cost, so calibration is not uniformly beneficial. All 793 lazy reply
calls equal counted clarification exchanges.

\subsection{Semantic prior and robustness extension}

A frozen training-only object--location prior identifies the true member in
71/85 unseen pairs (83.5\%, Wilson CI [74.2,89.9]) and 69/77 seen pairs
(89.6\%, [80.8,94.6]). These checkpoint-level rates show ranking information,
not calibrated probabilities. Construction symmetry blocks the generator decoder
while retaining semantic information that a .5 prior discards.

On 340 unseen scenarios, Semantic GAVA's deltas in interaction cost and declared
joint cost versus uniform GAVA are $-0.490$ (CI [$-0.609,-0.363$]) and $-0.420$
([$-0.579,-0.218$]). Against always verify, the interaction-cost delta is
$-0.563$ (CI [$-0.659,-0.469$]); the declared-joint-cost delta is $-0.493$
(CI [$-0.638,-0.305$]). All four Holm-adjusted $p<.001$. It makes 4/340
factual errors (98.8\% descriptive accuracy)
while completing every task without invalid actions. Its online responses,
trajectories, dialogue, and costs match primitive-arm replay in all 340 cases.

We then froze the same policy, costs, eight methods, and six-test Holm family
before a full run on the 77 non-overlapping seen checkpoints. Across 308
scenarios, Semantic GAVA again lowers interaction and declared joint cost versus uniform
GAVA by $-0.595$ and $-0.517$ (CIs [$-0.726,-0.459$] and
[$-0.714,-0.252$]). Versus always verify, the corresponding reductions are
$-0.655$ and $-0.577$ ([$-0.773,-0.537$] and [$-0.765,-0.321$]); all four
adjusted $p=.00030$. Against an identical-prior
fixed Bayes rule, its delta in declared joint cost is $-0.639$ ([$-1.275,-0.166$], adjusted
$p=.0169$) with 4 versus 32 factual errors. Against an identical-prior,
calibrated no-VOI policy, however, the delta in declared joint cost is $-0.208$
([$-0.630,0.224$], $p=.359$; 63/77 ties). This establishes neither superiority
nor equivalence. Semantic GAVA's 4/308 factual errors give 98.7\% descriptive
accuracy. Online and primitive replay match in 308/308 cases; all methods again
achieve complete task success and zero invalid actions.

The cost trade-off is directly interpretable: four errors at loss 6 add
$24/340=0.071$ unseen and $24/308=0.078$ seen to mean interaction cost,
explaining the smaller joint-cost gains after rounding. These are arithmetic
consequences of the reported counts, not additional outcomes.
The earlier calibration/loss sweeps remain synthetic sensitivity analyses.
Static wording invariance is an implementation check, not evidence about human
language.

\section{Discussion}

\subsection{What the comparisons establish}

The interface separates using a claim from gathering evidence. Accept changes
search preference; reject withholds it; verify inspects the world; clarify asks
the speaker. For HRI, this makes environmental checking and speaker consultation
distinct, costed choices before using factual advice. Whether people understand,
prefer, or trust these responses requires human evaluation.

Coverage matters: without local negative evidence, every resolved response is
correct, yet no false claim is rejected. Full GAVA resolves both twins through
inspection. Non-resolution does not establish repeated questioning or stalling:
the static harness stops at a clarification request. Online runs measure the
actions and executed replies that connect response choices to planner behavior.

The experiments separate three questions. Static ablations establish that
complete local evidence resolves factual claims. Semantic priors then reduce
interaction and joint cost against uniform GAVA and always verify in both
cohorts. Finally, the full policy lowers joint cost against an identical-prior
fixed rule in the seen replication: prior-only commitment leaves avoidable
declared loss.

Whether inspection adds value beyond calibrated clarification remains open:
63/77 checkpoints tie and the matched CI spans zero. The evidence supports the
configurable interface and its measured cost--accuracy trade-off, not a general
advantage for environmental VOI. This matched baseline and the perfect-speaker
threshold tie define the limits of the contribution.

\subsection{Pathway to physical deployment}

The claim and arbitration interfaces could connect navigation, manipulation,
and a perception-backed belief store. Physical deployment would require
calibrated sensing, skill-success monitoring, collision and force constraints,
irreversibility-aware loss, and recovery policies. These requirements describe
an implementation pathway; transfer beyond the present planner and simulation
remains untested.

\subsection{Limitations}

The benchmark uses templated English, one normalized location claim, and two
early intervention timings. Complete symbolic listings bypass uncertain
perception. Speech recognition, ambiguous reference, deceptive intent,
asynchronous intervention, and social cues are absent. False twins need not
resemble natural mistakes. Navigation abstracts geometry and makes checking cheap.

Online clarification supplies one controlled reply, sometimes a correct
replacement location, from an isolated service. Reliability levels and
action/error exchange rates are synthetic, not human estimates. ``Calibrated''
policies receive speaker parameters rather than learning them.
Equation~\eqref{eq:clarification} follows resolved replies instead of optimally
combining each answer with its prior. Dialogue cost and error loss do not
measure physical harm.

The semantic prior is a smoothed, pair-conditioned training estimate. Ranking
accuracy does not validate its probabilities or transfer to unpaired claims,
changed correction frequencies, or shifted locations. Multi-claim behavior is
covered only by software tests. Incomplete probes are descriptors, not noisy
trajectories; the one-step rule is not a full belief-space planner.

The planner is deterministic and hand engineered, using training counts.
The seen replication is prospective for new policy outcomes, not an untouched
domain: static results and aggregate prior predictability were already known.
Every method completes every task, and semantic GAVA's four errors per cohort
are recoverable. The study measures checking cost and declared decision loss,
not prevention of task failure or physical harm.

\section{Conclusion}

GAVA and the paired protocol make factual correction handling an executable
choice among acceptance, rejection, inspection, and clarification. A semantic
prior reduces interaction and declared joint cost against uniform GAVA and
always verify across exploratory and internally replicated cohorts; the full
policy also lowers joint cost against identical-prior fixed decisions. Four
errors per cohort expose the trade-off. Added value over matched calibrated
clarification remains inconclusive. The contribution is selective checking
under explicit evidence and cost assumptions. Broader correction distributions,
calibrated uncertainty, and human or physical evaluation are required for
claims about dialogue competence or safety.

\clearpage
\balance
\bibliographystyle{ACM-Reference-Format}
\bibliography{references}

\end{document}